\documentclass[letterpaper, 10 pt, conference]{ieeeconf}  

\IEEEoverridecommandlockouts                              

\usepackage{graphics} 
\usepackage{epsfig} 
\usepackage{mathptmx} 
\usepackage{times} 
\usepackage{amsmath} 
\usepackage{amssymb}  
\usepackage{graphicx}
\usepackage{hyperref}
\usepackage{graphics,graphicx,amssymb,amsmath,verbatim}
\usepackage{mathrsfs}
\usepackage{amsfonts}
\usepackage{epstopdf}
\usepackage{xcolor}
\usepackage{subfigure}
\usepackage{hyperref}
\usepackage{booktabs}

\usepackage{algorithm}
\usepackage{algorithmic}
\usepackage{subfigure}
\usepackage{makecell}
\usepackage{diagbox}
\usepackage{multirow} 

\newenvironment{proof of Proposition }[1][Proof of Proposition]{\noindent\textit{#1.} }{\ \rule{0.5em}{0.5em}}
\newenvironment{proof of Theorem 2}[1][Proof of Theorem 2]{\noindent\textit{#1.} }{\ \rule{0.5em}{0.5em}}
\newenvironment{proof of Theorem 3}[1][Proof of Theorem 3]{\noindent\textit{#1.} }{\ \rule{0.5em}{0.5em}}
\newenvironment{proof of Theorem 4}[1][Proof of
Theorem 4]{\noindent\textit{#1.} }{\ \rule{0.5em}{0.5em}}
\newenvironment{proof of Theorem 5}[1][Proof of Theorem 5]{\noindent\textit{#1.} }{\ \rule{0.5em}{0.5em}}

\title{\LARGE \bf
Single-Microphone-Based Sound Source Localization for Mobile Robots in Reverberant Environments
}

\author{Jiang Wang\textsuperscript{$\dagger$}, Runwu Shi\textsuperscript{$\dagger$}, Benjamin Yen, He Kong, and Kazuhiro Nakadai,~\IEEEmembership{Fellow,~IEEE} 
\thanks{Jiang Wang\textsuperscript{$\dagger$} and Runwu Shi\textsuperscript{$\dagger$} contributed equally to this work. This work was supported by JST BOOST, Grant No. JPMJBS2430. Jiang Wang (corresponding author), Runwu Shi, Benjamin Yen, and Kazuhiro Nakadai are with the Department of Systems and Control Engineering, Institute of Science Tokyo (formerly Tokyo Institute of Technology), Tokyo, Japan. Emails: 
\{wangjiang,shirunwu,benjamin,nakadai\}@ra.sc.eng.isct.ac.jp.
He Kong is with the School of Automation and Intelligent Manufacturing, Southern University of Science and Technology, Shenzhen, China. Email: kongh@sustech.edu.cn.} 
	}

\begin{document}

\maketitle
\thispagestyle{empty}
\pagestyle{empty}

\begin{abstract}
Accurately estimating sound source positions is crucial for robot audition. However, existing sound source localization methods typically rely on a microphone array with at least two spatially preconfigured microphones. This requirement hinders the applicability of microphone-based robot audition systems and technologies. To alleviate these challenges, we propose an online sound source localization method that uses a single microphone mounted on a mobile robot in reverberant environments. Specifically, we develop a lightweight neural network model with only 43k parameters to perform real-time distance estimation by extracting temporal information from reverberant signals. The estimated distances are then processed using an extended Kalman filter to achieve online sound source localization. To the best of our knowledge, this is the first work to achieve online sound source localization using a single microphone on a moving robot, a gap that we aim to fill in this work. Extensive experiments demonstrate the effectiveness and merits of our approach. To benefit the broader research community, we have open-sourced our code at https://github.com/JiangWAV/single-mic-SSL.
\end{abstract}

\section{Introduction}
Sound source localization (SSL) is a critical component of robot audition, enabling robots to accurately estimate the positions of sound sources and perform subsequent tasks \cite{Nakadai15,Rascon2017}. For instance, SSL allows robots to locate and interact with human speakers \cite{Younes2023}, conduct search and rescue missions in visually occluded environments \cite{AnTRO,YIN2024}, and perform acoustic scene mapping in unknown environments \cite{Fu2024}. These SSL methods rely on multiple microphones with a known geometric configuration, such as binaural setup or microphone array \cite{Rascon2017,jasa22}. By leveraging inter-channel phase difference (IPD), and inter-channel intensity differences (IID) from it, these methods extract spatial cues to estimate sound source positions.

However, deploying multiple microphones imposes constraints on space and portability, making them impractical for miniature robots with strict size limitations. Even for larger robotic platforms, having a microphone array onboard can be cumbersome. Furthermore, the multi-sensor nature of microphone arrays means that their performance depends on precise synchronization and calibration \cite{Kong2021,Jiang24}. If individual microphones malfunction, or introduce significant deviations, the overall system’s accuracy may be compromised.

In contrast, SSL with a single microphone offers a promising alternative. While prior research has explored single-microphone SSL, most existing methods simplify the problem to only sound source direction-of-arrival (DOA) estimation \cite{Takashima09,Andrew09,Youssef24}. This limitation poses challenges for applications such as robot audio-visual navigation, acoustic tracking, and scene analysis, where accurate spatial localization is essential. In addition, some single-microphone SSL methods attempt to compensate for the lack of spatial cues by integrating artificial pinna to enable single-ear localization \cite{Andrew09,Youssef24,Takashima10}. However, artificial pinna are uncommon in most robots and introduce additional constraints due to their physical size. 
\begin{figure}[t]
\centering 
{\includegraphics[width=0.9\columnwidth]{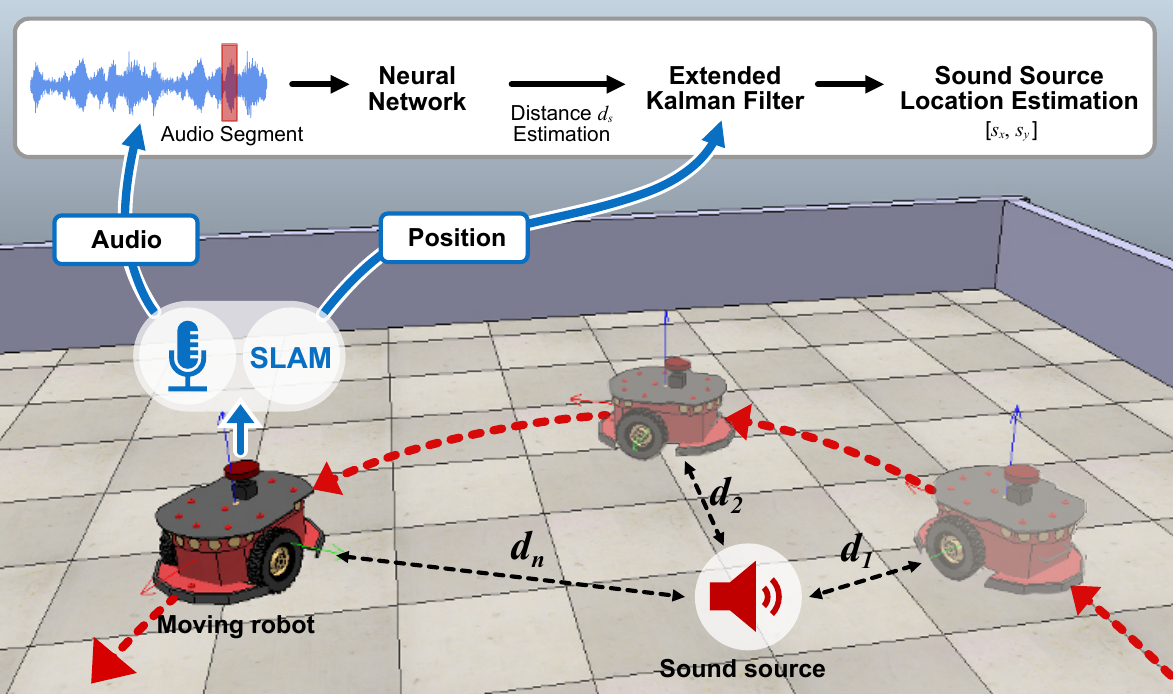}
}
\caption{Problem setup and sound source localization framework using a single microphone on a moving robot.}
\label{ssl} 
\vspace{-1em}
\end{figure}

To overcome the aforementioned limitations, leveraging room reverberation for single-microphone sound source localization is a viable approach \cite{shi2024distance}. The room impulse response (RIR) can be regarded as the transfer function between the sound source and the receiver, characterizing how sound propagates and interacts with the surrounding environment \cite{liu2025}. Moreover, this transfer property enables sound source distance estimation, as the direct-to-reverberant ratio (DRR) decreases with increasing distance between the source and the microphone \cite{patterson2022distance}. This observation motivates us to adopt distance as an intermediate variable for SSL in reverberant environments. As the robot moves, the change in its relative distance to the sound source can be used to infer the location of the sound source. 

In this paper, we propose a method for online sound source localization in reverberant environments using only a single microphone mounted on a moving robot, as illustrated in Fig. \ref{ssl}. Specifically, our method is made of the following few key steps. First, we develop a lightweight neural network model for real-time distance estimation, consisting of only 43k parameters. This model employs a subband processing strategy and operates on extremely short audio segments (0.2s in our experiments). To capture temporal information in reverberant signals, we introduce Filter-Attention (FA) blocks, which integrate multi-scale filters and the self-attention mechanism. The estimated distances are then processed by an extended Kalman filter (EKF) to perform online SSL. Extensive simulations and real-world experiments validate the effectiveness and merits of our approach.

\section{Related Work}
The field of SSL has been extensively studied, with a vast body of literature available. In this section, we focus specifically on research related to SSL using a single-channel audio data rather than providing a comprehensive review of all SSL methods.


\subsection{Monaural SSL}
Monaural SSL methods rely on the structure of an artificial pinna, which functions as filters that shape the sound transfer function from the source to the microphone. This filtering effect allows sounds from different directions to be distinguished. For example, in \cite{Andrew09}, a hidden Markov model is trained to represent the direction-dependent transfer function of the pinna, enabling the estimation of the sound source direction of arrival. In \cite{Youssef24}, the second-order derivatives of the spectra for different source directions are extracted and used as input features for a multi-layer perception to estimate sound source direction. \cite{Takashima10} employs a parabolic reflecting plate to create distinct acoustic transfer functions for target and non-target directions, facilitating sound source direction estimation. However, these methods rely on robots equipped with artificial pinna, such as humanoid robots, which are uncommon in most robot platforms.

\subsection{Single-Microphone SSL via Acoustic Reflection}
Unlike the aforementioned methods, single-microphone source localization techniques leverage acoustic reflection to infer spatial information about the source without relying on specialized structures. In \cite{Takashima09}, a Gaussian mixture model is employed to represent clean speech features and estimate the sound transfer function and speaker direction from reverberant speech using maximum likelihood estimation. Building on the work of \cite{Takashima09}, \cite{Takashima13} introduces a multikernel learning-based cepstral feature weighting method that enhances the direction of arrival estimation by selectively reweighting cepstral dimensions of the acoustic transfer function. \cite{Guo22} proposes two approaches, the composite reverberant speech model and the direct training reverberant speech model, to estimate the room sound transfer function and the speaker’s direction. Additionally, \cite{Vetterli14} associate recorded reverberation with the reflective walls that produced them, formulating a least squares problem under the s-stress criterion to estimate the sound source’s distance from the microphone. \cite{Neri24} utilizes a convolutional recurrent neural network with an attention module to infer the source-to-microphone distance in reverberant environments directly from the audio signal.

Our proposed method is closely related to previous studies (e.g., \cite{Vetterli14,Neri24}), which also infer sound source distance from sound reflections. However, these methods cannot estimate the precise source positions. Moreover, real-time performance was not a consideration in these studies. In contrast, we propose an online sound source localization method using a single microphone mounted on a mobile robot in reverberant environments.  Compared to \cite{Neri24}, our model is significantly more efficient, requiring only one-tenth of the parameters, supporting more flexible audio inputs, and enabling faster inference, making it more suitable for deployment on resource-constrained robot platforms. 

\section{\label{3D}Proposed method}
\subsection{Setup and Problem Statement}
We consider a two-dimensional indoor environment consisting of a robot and a stationary sound source, where the robot is equipped with a single microphone with known positions on the robot. As shown in Fig. \ref{ssl}, the robot is assumed to have a local frame attached to it, and the global frame is chosen to coincide with the robot frame at time instance $k=0$, {denoted as} $\left\{ \mathbf{G}\right\}$. In the process of sound source localization, the sound source remains stationary, and the mobile robot accurately estimates the sound source position in the global frame by taking measurements under different positions.

At any time instance $k$, where $k=1,\ldots,K$, the robot pose $\mathbf{r}_k$ is represented in the global frame by its position $\mathbf{p}_k=\left[x_{k};y_{k}\right]$ and yaw angle $\theta_k$ as  
\begin{equation}
\mathbf{r}_{k}=\left[\mathbf{p}_{k};\theta_{k}\right].
\end{equation}
We also denote the microphone position in the global frame as $\mathbf{m}_{k}$, which can be computed as follows:
\begin{equation}
\mathbf{m}_{k}=\mathbf{R}(\theta_{k})\mathbf{p}_{M}+\mathbf{p}_{k},
\end{equation}
where $\mathbf{R}\left(\theta_{k}\right)$ is the rotation matrix of the global frame to robot frame with $\theta_{k}$, and $\mathbf{p}_{M}$ is the microphone position on the robot. Thus, the stationary sound source position $\mathbf{s}$ to be identified in the global frame can be represented as
\begin{equation}
\mathbf{s}=\left[s_x;s_y\right].
\end{equation}
As such, the model of the robot sound source localization system via a single microphone is defined as:
\begin{equation}
\begin{aligned}
 \mathbf{a}_k= &\text{ }\mathbf{f}\left(\mathbf{m}_{k-1},\mathbf{u}_{k},\mathbf{s}\right)\\
\mathbf{z}_k=&\text{ }\mathbf{d}\left(\mathbf{m}_{k}, \mathbf{s}\right)+\mathbf{w}_{k}
\end{aligned}\label{eq:system}
\end{equation}
 where $\mathbf{a}_{k}$, $\mathbf{u}_{k}$, and $\mathbf{z}_{k}$ stand for the system states, the control input, and the sensor measurements, at the $k$-th sampling instant, respectively; $\mathbf{w}_{k}\sim\left(\mathbf{0},\mathbf{W}_{k}\right)$ is a Gaussian-distributed observation noise vector, with known covariance $\mathbf{W}_{k}$;  $\mathbf{f}(\cdot)$ and $\mathbf{d}(\cdot)$ represent the state transition function and the distance measurement function, respectively.

Given the above system model, our primary objective is 1) to extract sound source distance measurements from a single microphone audio data in reverberant environments and 2) use these measurements to accurately estimate the sound source position in the global frame.

\begin{figure*}[!htbp]
\centering 
\includegraphics[width=0.79\textwidth]{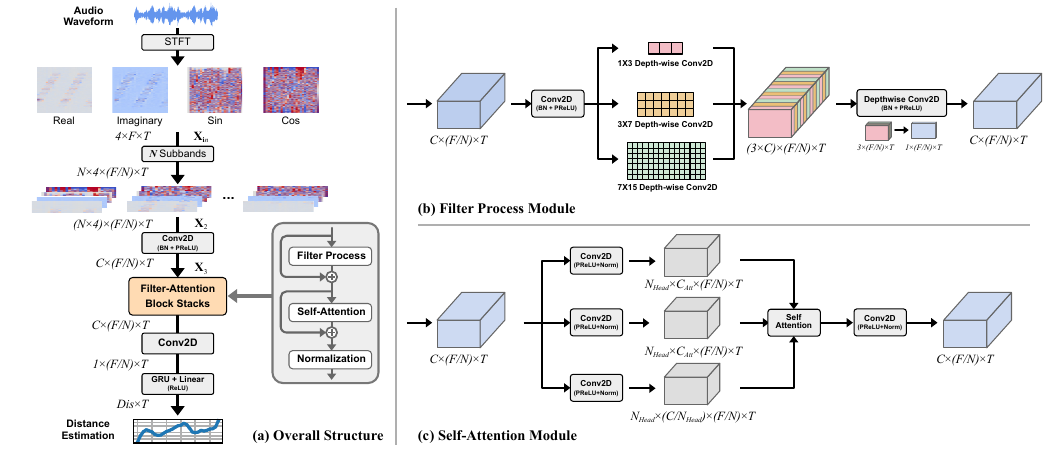}
\caption{(a) Overall structure of the neural network, (b) structure of Filter Process module, (c) structure of Self-Attention module. }
\label{network} 
\vspace{-1em}
\end{figure*}
\subsection{Lightweight Neural Network for Single-channel Sound Source  Distance Estimation}
To ensure suitability for resource-constrained robot platforms, we design a compact and lightweight neural network architecture that overlaps different subbands and extensively leverages depth-wise convolutions for filtering, providing high efficiency and accuracy. This approach achieves excellent performance while dramatically reducing model parameters. This model processes input audio of varying lengths end-to-end, as illustrated in Fig. \ref{network}.

{
To effectively capture the unique reverberation characteristics of each band, we adopt a subband processing strategy that decomposes the spectral signal into multiple subbands and extracts compact spectral features. 
The subband features are then processed by Filter-Attention (FA) blocks to capture time-frequency information at multiple resolutions from reverberant signals, which is crucial for sound source distance estimation. This multi-scale approach enables the model to extract both fine-grained and coarse features from reverberant signals more effectively. Simultaneous processing of all subbands reduces the feature map and decreases the overall model size. The depth-wise design keeps computational costs low, and a Self-Attention mechanism in the temporal dimension captures dependencies in the time domain. Finally, we used the gated recurrent unit (GRU) and linear layer combined with a ReLU activation function to output the estimated value by frame, and thus, the model allows the inputs of different lengths.}

Specifically, the network first converts the input reverberant audio $\mathbf{x} \in \mathbb{R}^{1 \times L} $ into its complex short-time Fourier transform (STFT) representation $\mathbf{X} \in \mathbb{C}^{F \times T}$, where $T$ is
the number of time frames, $F$ is the number of frequency bins,
and $L$ is the number of samples. Then, the phase information of the complex spectrogram is encoded in a smoother form using $\sin(\cdot)$ and $\cos(\cdot)$ functions to alleviate discontinuities within raw phase angles \cite{Siuzdak24}. Therefore, the input features are formed by concatenating the real and imaginary components of the complex spectrogram with the sine and cosine components of the phase, denoted as $\mathbf{X}_{in} = [\mathrm{Re}(\mathbf{X}), \; \mathrm{Im}(\mathbf{X}), \; \sin(\angle \mathbf{X}), \; \cos(\angle \mathbf{X})] \in \mathbb{R}^{4 \times F \times T}$, and are then passed to the following processing blocks. 
\subsubsection{\textbf{Subband Processing}}
 This step divides the input features into multiple frequency bands, allowing the model to capture fine-grained spectral information. 
 We uniformly divide the frequency bands, with the input features $\mathbf{X}_{in}$ split into $N$ subbands, each having a bandwidth of $F/N$ Hz. This results in $\mathbf{X}_{2}\in \mathbb{R}^{(N \times 4) \times (F/N) \times T}$. These subband features are then processed by a 2D convolutional layer with the kernel size of $1 \times 1$, followed by a batch normalization layer with a parametric rectified linear unit (PReLU), resulting in $C$ channel features $\mathbf{X}_{3}\in \mathbb{R}^{C \times (F/N) \times T}$ for subsequent processing. This convolutional layer can be viewed as a fusion of information from different frequency bands, while effectively reducing the size of the feature map to improve computational efficiency. 

\subsubsection{\textbf{Filter-Attention Block}}
The fused subband acoustic features $\mathbf{X}_{3}$ are fed into repeated FA blocks for further feature extraction, each FA block consists of a Filter Process module, a Self-Attention module, and channel normalization, as shown in Fig. \ref{network}. The Filter Process module contains several rectangular filters of different sizes, aiming to extract acoustic features at different time and frequency scales. Firstly, a 2D convolutional layer with a $1 \times 1$ kernel reorganizes the input feature map and outputs the same $C$ channels, followed by batch normalization and a PReLU activation function. Then, three depth-wise convolutional layers with kernel sizes $1 \times 3$, $3 \times 7$, and $7 \times 15$ are applied. Each convolution operates independently on its corresponding scale. The obtained features with the shape of $3 \times C \times (F/N) \times T$ are then processed by a depth-wise convolutional layer with a group size of $C$. In this configuration, the 3 channels corresponding to different kernel sizes for each of the $C$ groups are independently processed and fused into a single channel per group. The residual connection is adopted and as a result, the final output is a feature map of shape $C \times (F/N) \times T$. Utilizing depth-wise convolutional layers offers a significant benefit by reducing the parameter amount.

After that, the Self-Attention module is adopted \cite{wang2023tf, xu2024tiger}. Distinct $1 \times 1$ convolutional layers are utilized to learn the query matrix $Q \in \mathbb{R}^{ N_{Head} \times C_{Att} \times (F/N) \times T} $, key matrix $H \in \mathbb{R}^{N_{Head} \times C_{Att} \times (F/N) \times T} $, and the value matrix $Q \in \mathbb{R}^{ N_{Head} \times (C/N_{Head}) \times (F/N) \times T} $. The attention is computed along the temporal dimension, and for the $h$-th head, its matrix is reshaped into a $(C_{Att} \times (F/N)) \times T$ form. The $h$-th head's attention is calculated as: 
\begin{equation}
\mathrm{Softmax}\!\left(\frac{Q_hH_h^T}{\sqrt{(C_{Att} \times F)_{\text{dim}}}}\right)V_h.
\end{equation}
Different heads are then concatenated and passed through a final Self-Attention layer to obtain the feature with the original input shape of $C \times (F/N) \times T$. The Self-Attention module incorporates a residual connection, and finally, channel normalization is applied to the output to ensure a stable feature distribution for stable training.
\subsubsection{\textbf{Model Implementation and Training}}
For the model training, we build a simulated dataset consisting of generated RIR using FRA-RIR \cite{luo2024fast}, where each sample is labeled with its corresponding distance. The sweep signal is randomly convolved with the RIR signal. The model implementation is presented in Table~\ref{tab:model_parameters}.
We set the batch size to 84 and used the Adam optimizer with a learning rate of 0.001. If the loss doesn’t improve for 10 epochs, the learning rate is reduced to 80\% of its current value. The model is trained for 200 epochs using the mean squared error (MSE) as the loss function.
\begin{table}[t]
  \centering
  \caption{Model Implementation Parameters.}  
  \begin{tabular}{ll ll}
    \toprule
    Parameter & Value & Parameter & Value \\
    \midrule
    Frame length (STFT) & 32 ms  & Frame shift (STFT) & 8 ms \\
    Frequency bins      & 256    & Subbands (N)      & 16 \\
    FA blocks           & 4      & Channel (C)       & 32 \\
    $N_{Head}$               & 4      & $C_{Att}$              & 4 \\
    \bottomrule
  \end{tabular}
  \label{tab:model_parameters}
  \vspace{-1.5em}
\end{table}

\subsection{\label{ekf}Online Sound Source Localization}
To perform online sound source localization, this paper employs the widely used EKF to recursively estimate the sound source position online. Denote $\hat{\mathbf{s}}$ and $\hat{\mathbf{P}}$ as the estimated mean and the variance of the sound source position, respectively. During SSL, the sound source is assumed to be static. The prediction and correction steps are
\begin{equation}\label{predict}
\mathbf{\hat{s}}_{k|k-1}=\mathbf{\hat{s}}_{k-1|k-1},\text{ }\mathbf{\hat{P}}_{k|k-1}=\hat{\mathbf{P}}_{k-1|k-1}
\end{equation}
and 
\begin{equation}\label{update}
\begin{aligned}\hat{\mathbf{s}}_{k|k}= & \mathbf{\hat{s}}_{k|k-1}+\mathbf{K}_{k}\left(\mathbf{z}_{k}-\mathbf{d}(\mathbf{m}_{k}, \hat{\textbf{s}}_{k|k-1})\right)\\
\mathbf{\hat{P}}_{k|k}= & \left(\mathbf{I}-\mathbf{K}_{k}\mathbf{G}_k\right)\mathbf{\hat{P}}_{k|k-1}
\end{aligned},
\end{equation}
respectively, where $\mathbf{z}_{k}$ is the distance estimation using the neural network and $\mathbf{d}(\mathbf{m}_{k}, \hat{\textbf{s}}_{k|k-1})$ can be calculated as
\begin{equation}
\mathbf{d}(\mathbf{m}_{k}, \hat{\textbf{s}}_{k|k-1}) = \left\Vert \mathbf{\hat{s}}_{k|k-1}-\mathbf{m}_{k}\right\Vert_{2},
\end{equation}
and $\mathbf{K}_{k}$ is called the Kalman gain and is calculated as
\begin{equation}
\mathbf{K}_{k}=\mathbf{\hat{P}}_{k|k-1}\mathbf{G}_k^{\top}\left(\mathbf{G}_k\mathbf{\hat{P}}_{k|k-1}\mathbf{G}_k^{\top}+\mathbf{W}_{k}\right)^{-1},
\end{equation}
where $\mathbf{G}_k$ is the Jacobian matrix of the observation model in (\ref{eq:system}), and can be calculated as
\begin{equation}
\mathbf{G}_k=\left(\frac{\partial\mathbf{d}}{\mathbf{\partial\mathbf{s}}}\right)\vert_{\mathbf{m}_{k},\mathbf{\hat{s}}_{k|k-1}}\\
= \frac{(\mathbf{\hat{s}}_{k|k-1}-\mathbf{m}_{k})}{\left\Vert \mathbf{\hat{s}}_{k|k-1}-\mathbf{m}_{k}\right\Vert _{2}}.
\end{equation}

\begin{table}
    \centering 
    \renewcommand\arraystretch{1.2}
    \caption{\label{tab:1}Simulations Dataset Parameters}
    \begin{tabular}{p{0.8cm} p{0.8cm} p{1.2cm} p{1.2cm} p{1.3cm} p{0.8cm}}
    \toprule
    Dataset & Room & RT60 (s) & Mic num & Src per mic & Total\\
    \midrule
    Sim1 & 1  & 0.6  & 900 & 50 & 45,000 
    \\
    Sim2 & 10 & 0.4 - 0.75 & 100 & 50 & 50,000 
    \\
    Sim3 & 100 & 0.4 - 0.75 & 100 & 50 & 500,000  
    \\
    Sim4 & 90 & 0.4 - 0.75 & 100 & 50 & 450,000  
    \\
\toprule
\end{tabular}
\end{table}
\begin{figure}[t]
\centering 
{\includegraphics[width=0.65\columnwidth]{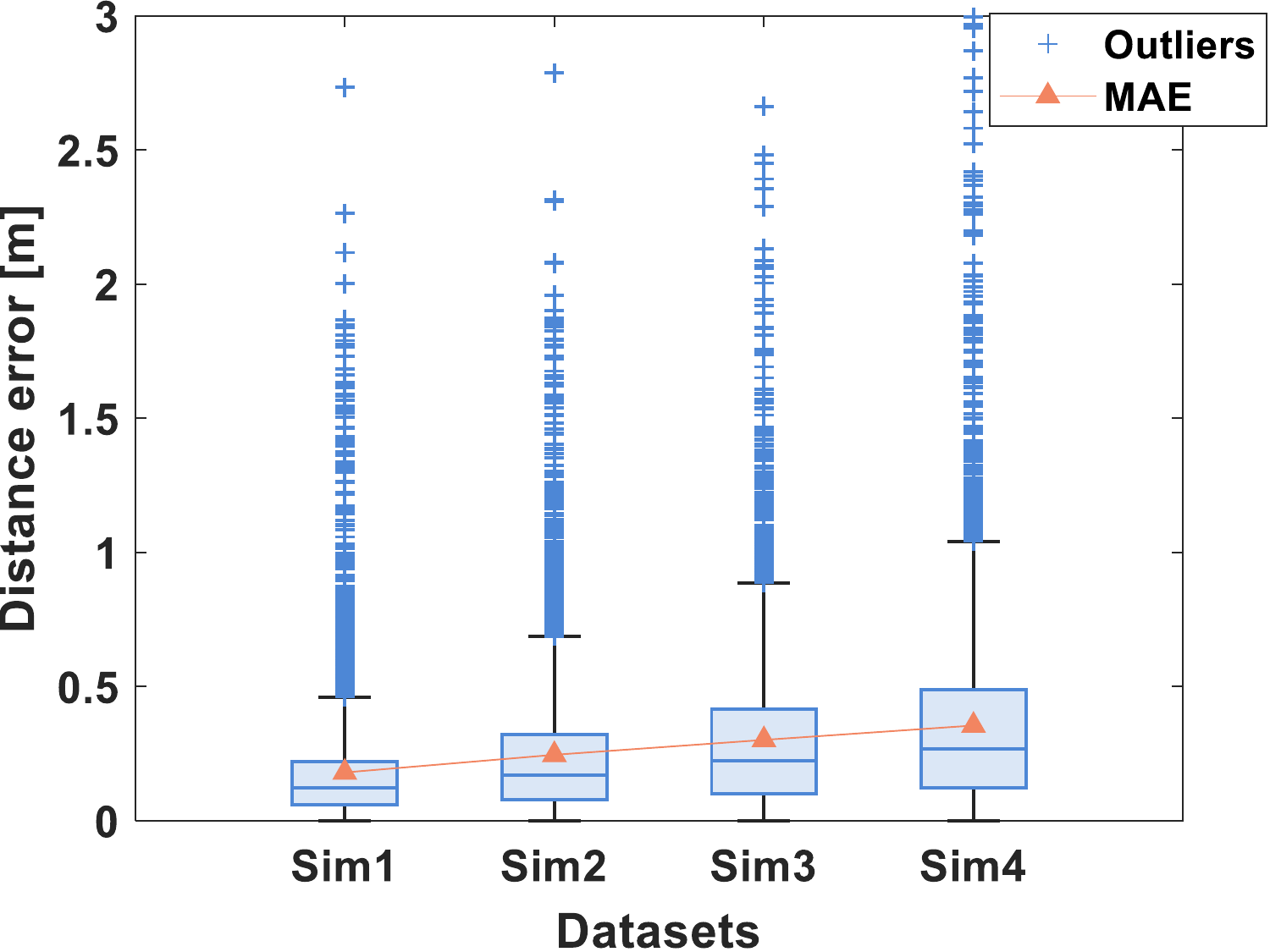}
}
\caption{Error distributions between the estimated sound source distance and true values for different datasets.}
\label{room} 
\vspace{-1.5em}
\end{figure}

\begin{table}[t]
    \begin{center}
    \renewcommand\arraystretch{1.2}
    \caption{\label{tab:2}Comparasion of Sound Source Distance Estimation With Other Method On Sim1 (Bold Means Better)}
    \centering 
    \begin{tabular}{p{1.6cm}p{1.4cm}p{1.4cm}p{1.4cm}}
    \toprule
    Model &	MAE (m) & Params & MACs \\
    \midrule
    Proposed & 0.179 & \textbf{43k} & \textbf{17.4M} \\
    SeldNet \cite{Neri24} & \textbf{0.134} & 649k & 93.6M \\
\toprule
\end{tabular}
\end{center}
\vspace{-2em}
\end{table}

\section{\label{simu}Simulation Experiments for Sound Source Distance Estimation}
To evaluate the performance of the proposed single-microphone sound source localization method, we conducted a series of experiments in simulations. First, we evaluated the model's performance across different room configurations to examine its generalization capability. Then, we benchmarked our method against the state-of-the-art \cite{Neri24} in single-microphone distance estimation, evaluating estimation accuracy, model complexity, and computational cost.

To simulate reverberation in rooms, we first create three datasets: Sim1, Sim2, and Sim3, each containing RIR signals using different numbers of rooms (1, 10, and 100, respectively). Table \ref{tab:1} summarizes the dataset configuration, where Mic num refers to the number of microphones in each room, and Src per mic denotes the number of sound sources corresponding to each microphone. The three datasets are used for training and testing on samples collected from all available rooms, with varying microphone and source positions. The splits for training, validation, and testing are set at 0.8:0.05:0.15. To further evaluate generalization to unseen rooms, we re-partition the Sim3 dataset to create the Sim4 dataset, where 90\% of the rooms are reserved for training and the remaining rooms for testing. The room dimensions are randomly sampled within the range of 5.4m × 6.4m × 2.5m to 6.4m × 7.4m × 3.5m. The chirp signal is synthesized at 0-8000 Hz with a period of 0.1s. 

We evaluated the accuracy of distance estimation using the mean absolute error (MAE). As presented in Fig. \ref{room}, the results represent the distribution between the estimated and ground truth distances across the four test sets. 
It indicates that the proposed method accurately estimates sound source to microphone distances across datasets of varying scales, yielding comparable performance. While a slight performance drop as a large number of rooms are included, these results show the robustness and generalization capability of our approach.

Next, we compare the proposed method with SeldNet \cite{Neri24}, the state-of-the-art approach for estimating sound source distance using a single microphone. Using the same training configuration and dataset (Sim1), we trained and tested SeldNet to evaluate its performance under identical conditions. We evaluated the two models based on distance estimation accuracy, parameter count, which reflects the model size, and the number of multiply-accumulate operations (MACs), which indicate computational complexity. The results summarized in Table \ref{tab:2} demonstrate that our method achieves comparable distance estimation performance while requiring fewer parameters and lower computational cost. This efficiency is largely attributed to our strategy of subband processing, which simultaneously processes all subbands, and the use of multi-scale depth-wise rectangular convolutions to capture time-frequency information at different resolutions.



\begin{figure}[t]
        \centering
	\begin{minipage}{0.8\linewidth}
		\centering
		\subfigure[]{\includegraphics[width=1\linewidth]{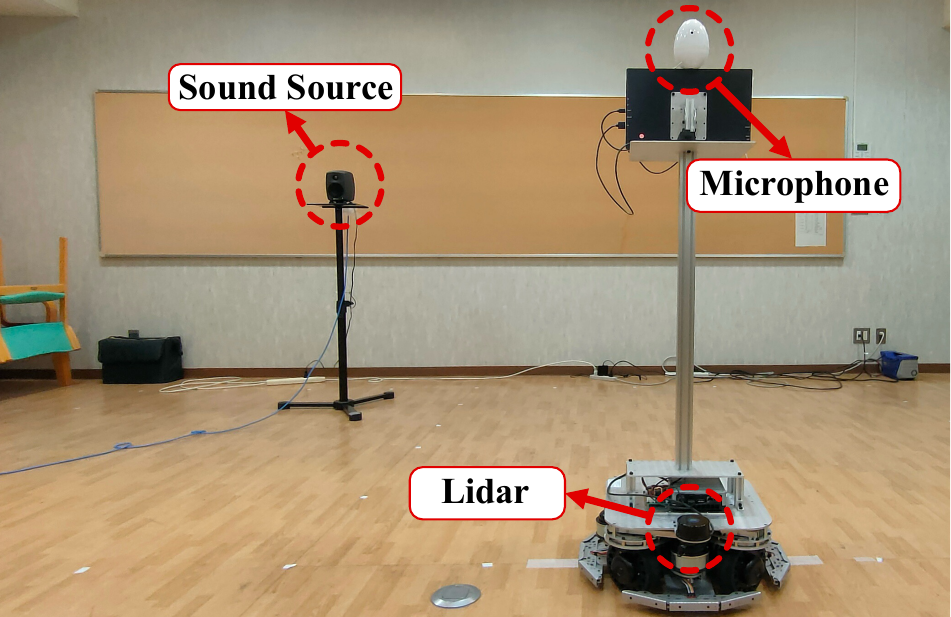}}
	\end{minipage}
	\centering
	\begin{minipage}{0.8\linewidth}
		\centering
		\subfigure[]{\includegraphics[width=1\linewidth]{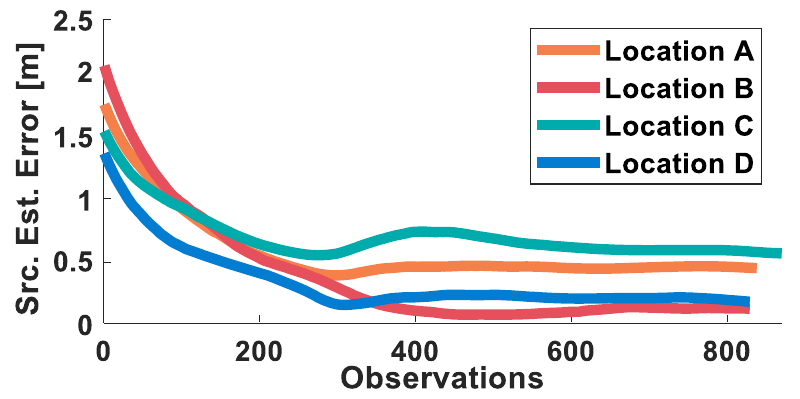}}
	\end{minipage}
	\begin{minipage}{0.9\linewidth}
		\centering
		\subfigure[]{\includegraphics[width=1\linewidth]{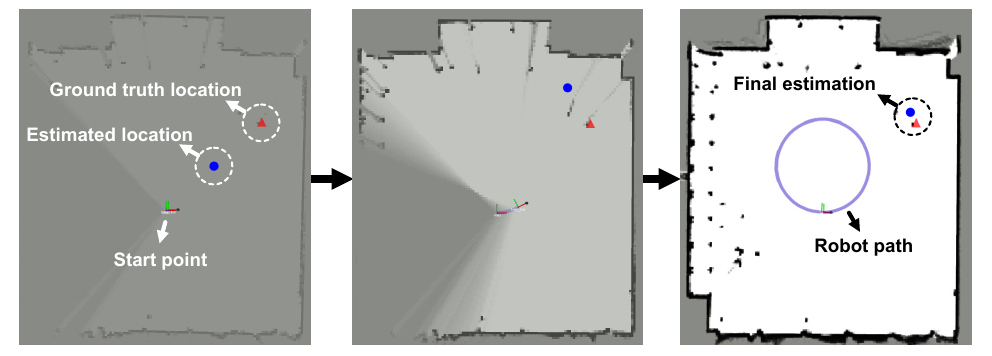}}
	\end{minipage}
\caption{Real-world sound source position estimation using a single microphone: experiments setup and results (a) Experiment setup, (b) The estimation error of the sound source positions, (c) The estimation results of the sound source position at different time instance.}
    \label{fig:exp_scene}
\vspace{-1.5em} 
\end{figure}

\section{\label{REAL}Real World Experiments for SSL}
\subsection{Experimental Setup}
To further evaluate and validate the proposed method, we conducted sound source localization experiments in a reverberant environment using a real robot platform. The experiments were performed on the Vstone 4WDS ROVER X40A robot, which was equipped with a computer with an AMD R5 3550H processor, a LiDAR, and a microphone array, as shown in Fig. \ref{fig:exp_scene}(a). The microphone array, TAMAGO-01, consists of eight microphones arranged in a circular pattern. Each microphone has a 24-bit sampling depth and a sampling rate of 16 kHz. For the experiments, we used data from only the first microphone of the array for sound localization. The LiDAR, YDLiDAR TG30, was used to provide the robot poses in real-time by Cartographer \cite{cartographer} in the robot operating system (ROS) at 0.05m resolution. The speaker continuously emits a chirp signal with a frequency range of 0 to 8000 Hz and a period of 0.1s.

To bridge the gap between simulated training and real-world deployment, we fine-tune the model originally trained on the Sim3 dataset, as described in Section IV. Specifically, we update the model for 100 epochs using the recorded audio from five known sound source positions in the target room (5.9 m × 6.9 m × 2.9 m), each labeled with its corresponding measured distance. After fine-tuning, we evaluate the model to infer the new positions of the sound source. During the sound source localization experiments, the robot initial position is viewed as the origin of the global frame, with its initial velocity along the x-axis and the y-axis extending to its left, forming a right-handed coordinate system. The robot moves along a circular trajectory and continuously receives audio streams in real-time and publishes 0.1s audio segments via ROS topics. The default length for audio segment processing by neural
network is 0.2s.

In real-world experiments, we conducted two sets of experiments to investigate the effect of the sound source distance from the wall and the length of audio segments used for processing on SSL performance. In the first set of experiments, the sound source was placed at four positions in the global frame (unit: meters): (2, 2.5), (2.5, 2.5), (1.5, 2.5), and (2, 2), labeled as Locations A, B, C, and D, respectively. A larger x-coordinate indicates a position closer to the wall. In the second set of experiments, the sound source was placed at (2, 3), and we recorded five datasets, processing each with audio segment lengths of 0.2s, 0.4s, 0.6s, and 0.8s, labeled as Test A, B, C, D, and E, respectively. For all experiments, the EKF was initialized with the sound source at (1,1) and the covariance matrix as the identity matrix.

\begin{table}[t]
    \begin{center}
    \renewcommand\arraystretch{1.2}
    \caption{\label{tab:4}Results of SSL Under Varying Audio Input Length (Bold Means Better)(Unit: \text{m})}
    \centering 
    \scalebox{0.9}{\begin{tabular}{p{1.5cm}p{1.2cm}p{1.2cm}p{1.2cm}p{1.2cm}}
    \toprule
    \multicolumn{1}{l}{\multirow{2}{0.5cm}{Datasets}} &\multicolumn{4}{c}{Input Length} \\ 
            \cmidrule(lr){2-5}
     & 0.2s & 0.4s & 0.6s & 0.8s\\
    \midrule
    Test A & 0.786 & 0.575 & 0.576 & \textbf{0.573}\\
    Test B & 0.787 & \textbf{0.589} & 0.593 & 0.594\\
    Test C & 0.780 & \textbf{0.584} & 0.588 & 0.588\\
    Test D & 0.790 & \textbf{0.589} & 0.591 & \textbf{0.589}\\
    Test E & 0.788 & 0.588 & 0.588 & \textbf{0.586}\\
    MAE & 0.786 & \textbf{0.585} & 0.587 & 0.586\\
\toprule
\end{tabular}}
\end{center}
\vspace{-1.5em}
\end{table}

\subsection{SSL Results}
Fig. \ref{fig:exp_scene}(b) shows the SSL results for different distances of the sound source from the wall. It can be observed that the robot successfully ensures the convergence of the estimator across all experiments at varying distances, confirming the effectiveness of the proposed method. Notably, when the sound source is closest to the wall at Location B, the localization results are more accurate. However, at Location C, where the sound source is farthest from the wall, the localization results are less accurate. These results indicate that the position of reflective surfaces significantly affects the accuracy of sound source localization. Fig. \ref{fig:exp_scene}(c) illustrates a representative result, showing the sound localization process as the robot moves. 

Table \ref{tab:4} shows the effect of varying audio segment lengths on SSL performance. It is evident that using longer audio segments significantly improves localization accuracy compared to the 0.2s segment. However, increasing the segment length beyond 0.4s does not result in a noticeable improvement. This is due to the fact that the sound source emits rapid chirp signals, contributing less information over longer time scales.

\section{\label{CONCLUSION}CONCLUSION}
In this study, we propose an online sound source localization method that relies on a single microphone mounted on a moving robot in reverberant environments. A lightweight neural network model with only 43k parameters is developed to extract temporal information from reverberant signals for real-time distance estimation. The estimated distances are then processed using an extended Kalman filter to achieve online source localization. 
Extensive experiments confirm that our method enables localization using only a single microphone mounted on a moving robot. In the future, we aim to extend to more challenging 3D scenarios involving multiple and diverse types of sound sources and more complex environments.

\end{document}